\documentclass[letterpaper, 10 pt, conference]{ieeeconf}  

\IEEEoverridecommandlockouts                              

\usepackage{graphics} 
\usepackage{epsfig} 
\usepackage{mathptmx} 
\usepackage{times} 
\usepackage{amsmath} 
\usepackage{amssymb}  

\usepackage[utf8]{inputenc}
\usepackage{cite}
\usepackage{amsmath,amssymb,amsfonts}
\usepackage{algorithmic}
\usepackage{graphicx}
\usepackage{textcomp}
\usepackage{xcolor}
\usepackage{bm}
\usepackage{multirow}
\usepackage{siunitx}
\usepackage{tikz}
\usepackage[switch]{lineno}  
\usepackage[hyphens]{url}  
\newcommand\tstrut{\rule{0pt}{2.7ex}}
\newcommand\bstrut{\rule[-1.0ex]{0pt}{0pt}}
\def\BibTeX{{\rm B\kern-.05em{\sc i\kern-.025em b}\kern-.08em
		T\kern-.1667em\lower.7ex\hbox{E}\kern-.125emX}}
\usepackage[caption=false, font=footnotesize]{subfig}

\usepackage{textcomp}
\newcommand\copyrighttext{%
	\footnotesize \textcopyright 2021 IEEE. Personal use of this material is permitted. Permission from IEEE must be obtained for all other uses, in any current or future media, including reprinting/republishing this material for advertising or promotional purposes, creating new collective works, for resale or redistribution to servers or lists, or reuse of any copyrighted component of this work in other works. DOI: {t.b.a}}
\newcommand\copyrightnotice{%
	\begin{tikzpicture}[remember picture,overlay]
	\node[anchor=south,yshift=10pt] at (current page.south) {\fbox{\parbox{\dimexpr\textwidth-\fboxsep-\fboxrule\relax}{\copyrighttext}}};
	\end{tikzpicture}%
}

\begin{document}

\title{\LARGE \bf
	Variational Autoencoder-Based Vehicle Trajectory Prediction with an Interpretable Latent Space}

\author{Marion Neumeier\textsuperscript{1}*, Andreas Tollk\"{u}hn\textsuperscript{2}, Thomas Berberich\textsuperscript{2} and Michael Botsch\textsuperscript{1}
	\thanks{*We appreciate the funding of this work by AUDI AG.}
	\thanks{$^{1}$ Technische Hochschule Ingolstadt, CARISSMA Institute of Automated Driving (C-IAD), 85049 Ingolstadt\newline
		{\tt\small \{marion.neumeier, michael.botsch\}@thi.de}}%
	\thanks{$^{2}$AUDI AG, 85057 Ingolstadt\newline
		{\tt\small \{andreas.tollkuehn, thomas.berberich\}@audi.de}}%
}

\maketitle
\copyrightnotice


\begin{abstract}
	This paper introduces the \textit{Descriptive Variational Autoencoder (DVAE)}, an unsupervised and end-to-end trainable neural network for predicting vehicle trajectories that provides partial interpretability. The novel approach is based on the architecture and objective of common variational autoencoders. By introducing expert knowledge within the decoder part of the autoencoder, the encoder learns to extract latent parameters that provide a graspable meaning in human terms. Such an interpretable latent space enables the validation by expert defined rule sets. The evaluation of the DVAE is performed using the publicly available highD dataset for highway traffic scenarios. In comparison to a conventional variational autoencoder with equivalent complexity, the proposed model provides a similar prediction accuracy but with the great advantage of having an interpretable latent space. For crucial decision making and assessing trustworthiness of a prediction this property is highly desirable. 
\end{abstract}

\section{Introduction}
One major drawback of deep learning models is the lack of interpretability. Although data-driven approaches achieve outstanding performance in various tasks, it is hard to trust their predictions due to lacking information on their internal logic. A potential risk is, for example, that datasets may comprise misleading correlations that a Neural Network (NN) starts to rely on erroneously during training. As a result, the prediction accuracy for the available data might be statistically satisfactory but the basis for the model's decision is highly incorrect. A widely known example for a falsely learned
conclusion is the \textit{Husky or Wolf} classifier.\cite{Ribeiro.2016} Detecting such a fallacy by just testing the network and examining the raw data is hard or extremely time-consuming. Therefore, currently a lot of scientific investigation is done to fathom the functionality and decision-making basis of NNs. The objective of this research area is to provide more insight about \textit{why} and \textit{how} Machine Learning (ML) based judgments are made. Thereby, literature differentiates between two different (but related) purposes: \textit{Explainability} and \textit{Interpretability}.\cite{Gilpin.2018} While explainability refers to post-hoc techniques for gaining comprehension of existing Artificial Intelligence (AI) models and/or their decisions, interpretability aims for an intrinsic understanding in human terms. In this work the latter is pursued.

\begin{figure}[t]
	\centering
	\includegraphics[width=0.7\columnwidth]{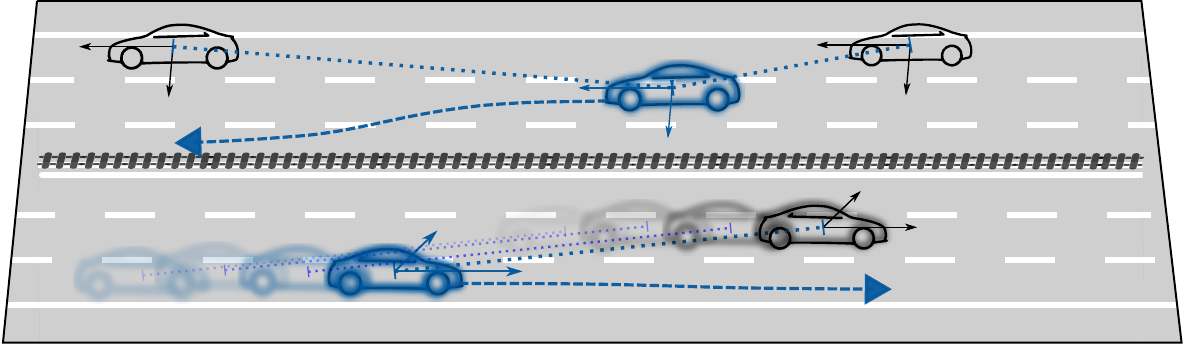} 
	\caption{A cooperative context, including spatial and temporal dimensions, describes the interaction of road users.}
	\label{fig2}
\end{figure}

Due to the great success of AI in solving real-life challenges, lots of research in the field of autonomous driving is currently also focusing on AI-based vehicle trajectory prediction. Predicting the intended motion of other traffic participants is a crucial task for safe driving of autonomous vehicles. Wrong predictions can cause fatal outcomes and therefore non-interpretable models cannot be trusted blindly. Data-driven models, however, manage to capture the high complexity of the \textit{cooperative context} within traffic scenarios. The cooperative context describes the situational interactions between traffic participants, as illustrated for example in Figure \ref{fig2}. For predicting trajectories, a lot of information lies within these interactions. The aim of this work is to introduce a NN, that is able to capture the cooperative context of a scenario and also to provide a partial intrinsic interpretability. Interpretability is a major aspect towards the usage and acceptance of ML for safety critical applications.

\textbf{Contribution.} The paper contributes towards establishing interpretability within NNs. Exemplarily, it tackles the challenge of how to design a long-term prediction model for vehicle trajectories that provides partial interpretability. The central aspects are:
\begin{itemize}
	\item A variational autoencoder-based architecture with an interpretable latent space is introduced. The conception enables an unsupervised method to generate interpretable latent spaces.
	\item The approach presents a way to easily validate ML models in safety critical applications, by using the interpretable latent space and physics-based rules.
	\item The proposed architecture is applied and tested for the task of vehicle motion prediction on highways. 
\end{itemize}
The content is organized as follows: Section II provides a review on the background and current state of the research fields. Section III introduces the concept of the proposed model. Section IV presents the performance of the approach for vehicle trajectory prediction and the comparison to baseline models. Finally, a conclusion regarding the architecture and results is drawn. In this work vectors are denoted as bold lowercase letters and matrices as bold capital letters.


\section{Related work}
The following section highlights relevant information regarding the paper's key topics and includes references to recent and related publications that inspire this work.
\subsection{Interpretability of ML}
In \cite{DoshiVelez.2017} the concept of interpretability is introduced as "\textit{The ability to explain or to present in understandable terms to a human.}" The aim of interpretability is to create a model with a comprehensible internal logic or interface. The most straightforward way to accomplish interpretability is to use interpretable-by-nature models like, e.\,g., decision trees.\cite{Carvalho.2019} These models do provide global comprehension on a modular horizon. Generally, existing approaches that provide interpretability can be distinguished by their scope and are categorized in \textit{model-agnostic vs. model-specific} and \textit{local vs. global} techniques.\cite{Hall.2019}
Interpretability is a domain-specific concept. It is an extremely subjective idea and, thus, not trivial to formalize.\cite{Carvalho.2019} The success of making a model interpretable is hinge on the background of each individual user due to different cognitive abilities and biases.\cite{Gilpin.2018} What is interpretable by a person with domain specific knowledge might not be interpretable by a non-expert. Therefore, an overall general notion cannot be established or would be inadequate.\cite{Rudin.2019}\cite{Carvalho.2019} In comparison to accuracy metrics, it is difficult to evaluate the aspects of interpretability in a fair manner. Due to this lack of a common evaluation criteria and the ambiguity in the meaning of interpretability regarding ML, benchmarking is difficult to carry out in practice. 

In literature several works can be found that claim their architecture to be at least partially interpretable. Wu \textit{et al.} \cite{Wu.2017} proposed an autoencoder-setup for reconstructing scenes with a similar strategy as the one presented in this work. They introduce an encoder-decoder network to learn an interpretable representation of scenes by implementing the decoder as a deterministic rendering function. The encoder extracts expressive information of a picture (e.\,g. number of objects) within the latent space and the decoder maps this naturally structured scene description into an image in order to recreate the input picture. Their approach, however, holds the drawback that the training process has to be semi-supervised at least. In comparison, the proposed model of this work does not require any labels, hence, can be trained unsupervised.

Based on the same idea but the objective of variational autoencoders, a comparable approach for education assessment was proposed in \cite{descVAE2019}. In this work, the decoder is replaced by a multidimensional logistic 2-parameter model, that allows to interpret the latent space.

\subsection{Motion Prediction Models}
Motion prediction of vehicles can be categorized in two main objectives: short-term predictions ($<$ \SI{1}{\second}) and long-term predictions ($\geq$ \SI{1}{\second}).\cite{Lefevre.2014} Conventional vehicle dynamic models, e.\,g., single-track models, do perform well in short-term predictions.\cite{Schubert.2008} Yet, when it comes to predicting trajectories within longer prediction periods their suitability is enormously limited.  A lot of research has been done in developing new techniques to improve prediction performance. An extensive survey on different approaches for motion prediction was done by Lef\`{e}vre \textit{et at.} \cite{Lefevre.2014}. While some methods try to improve model-based approaches, e.\,g. by introducing maneuver recognition modules \cite{Houenou.03.11.201307.11.2013}, data-driven approaches gain increasing importance.\cite{Gupta.2018}\cite{Lee.21.07.201726.07.2017}\cite{RohanChandra.2018}
Following the great success in many time series prediction tasks, \textit{Long Short-Term Memory (LSTM)} NNs are also widely used in predicting paths and trajectories of traffic participants.\cite{Altche.2018}\cite{Park.2018}\cite{Deo.2018}\cite{LstmflightTrajectory2018}\cite{ChengHaoLSTM2018} The pioneering work of Alahi \textit{et al.} \cite{Alahi.2016} on pedestrian trajectory prediction is based on LSTMs. Each trajectory in a scenario is the input to a separate LSTM, capturing the individual motion property. The pooled hidden-states of all LSTMs are shared among their neighbouring units in order to jointly reason about the future trajectories. Deo and Trivedi adapted this approach for vehicle predictions on highways.\cite{Deo.2018} They constitute a model that concludes the future trajectory of vehicles based on an autoencoder architecture. Each cell of a grid-based representation of a traffic scenario is encoded by a separate LSTM and then passed through convolutional layers. The latent representation is processed by a LSTM decoder to predict the trajectory. These data-driven approaches, however, do not provide insight on the internal logic. 
Hu \textit{et al.} \cite{Hu.2019} claimed to propose an interpretable prediction model, since he is "\textit{[...] able to explain each sampled data point and relate it to the underlying motion pattern}". In his work, a LSTM is used to extract important features of the recent trajectory and the driving intentions are determined by a Dynamic Time Warping Module.
Drawbacks of it are, that it is limited to only consider two interacting agents in any situation and reference trajectories need to be defined manually. 

\subsection{Autoencoder}
An Autoencoder (AE) is a NN that is trained to reproduce a model`s input to its output and, thus, the learning process is regarded as unsupervised.\cite{Goodfellow.2016} Instead of learning just to copy the input, AEs master to extract sensitive features, which enable to reconstruct the input. The architecture of any AE can be decomposed into two main parts: an \textit{encoder} and a \textit{decoder}. The encoder part performs the mapping of an input $\bm{x}$ to an internal representation $ \bm{z} \in \mathbb{R}^\mathcal{Z}$ represented by the function $ \bm{z} = f_{\boldsymbol{\theta}_e}(\bm{x}) $. This internal representation $ \bm{z}$ is usually referred to as the \textit{latent space}. The decoder computes a reproduction $ \bm{r} $ of the input by applying the function $ \bm{r} = g_{\boldsymbol{\theta}_d}(\bm{z}) $ onto the latent space representation $ \bm{z} $. Dissimilarity between the input data and the model`s reproduction is penalized and used as the learning objective. A frequently used loss function for evaluating dissimilarity is the Mean Squared Error (MSE)
\begin{equation}
\mathcal{L}_{\textrm{AE}}(\bm{x}) = || \bm{x} - g_{\boldsymbol{\theta}_d}(f_{\boldsymbol{\theta}_e}(\bm{x})) ||^2 = ||\bm{x} - g_{\boldsymbol{\theta}_d}(\bm{z}) || ^2 \, .
\label{eq:lossAE}
\end{equation}
Based on the reconstruction error of Eq.\,\ref{eq:lossAE} the AE parameters $ {\boldsymbol{\theta}_e} $ are ${\boldsymbol{\theta}_d}$ learned.

\textbf{Undercomplete Autoencoder.}
By setting up an encoder-decoder architecture, where at least one internal layer has a smaller dimension than the input space, the model can not just copy the input to the output. These realizations of AEs are called undercomplete autoencoders.\cite{Goodfellow.2016} Such architectures are forcing the AE to determine more efficient representations of the input data. The underlying idea is that this latent space holds (almost) the same information as the input but the information density is increased due to the reduced dimensionality.

\textbf{Variational Autoencoders.}
The idea of VAEs is to encode the input into a conditional distribution over the latent vector $ p(\bm{z} \vert \bm{x}) $ instead of a single latent vector $ \bm{z} $. The true distribution $ p(\bm{z} \vert \bm{x}) $, however, is intractable. Therefore, the goal of a VAE is to infer the true conditional density of latent variables $ p(\bm{z} \vert \bm{x}) $ through the stochastic encoder $ q_{\boldsymbol\phi}(\bm{z} \vert \bm{x}) $ given the observed data $ \bm{x} $. VAEs have been introduced by \cite{Kingma.b2014}. This paragraph summarizes their main aspects. 
Usually the (non-obligatory) key assumptions are made, that 
\begin{itemize}
	\item the prior $ p_{\boldsymbol\theta_1}(\bm{z})= \mathcal{N}(\bm{z}; \bm{0},\mathbf{I}) $ over the latent space $\bm{z}$ is a multivariate Gaussian distribution with zero mean and a diagonal covariance matrix, containing only ones in the diagonal, and 
	\item the proposal distribution $q_{\boldsymbol\phi}(\bm{z} \vert \bm{x})  = \mathcal{N}(\bm{z}; \bm{\mu}, \bm{\sigma}^2\mathbf{I})$ and the conditional probability $ p_{\boldsymbol{\theta}_2}(\bm{x}|\bm{z})$ are assumed to be a multivariate Gaussians with a diagonal covariance.
\end{itemize}
Due to the consideration of the latent features as distributions, the optimization objective of the VAE is carried out according to Eq.\,\ref{eq:lossVAE}.
\begin{align}
\begin{split}
\mathcal{L}_{\textrm{VAE}}(\bm{x}) = &-D_{\textrm{KL}}\big(q_{\boldsymbol{\phi}}(\bm{z}|\bm{x}) || p_{\boldsymbol{\theta}_1}(\bm{z})\big) \\
&+ \mathbb{E}_{q_{\boldsymbol{\phi}}(\bm{z}|\bm{x})} [\log p_{\boldsymbol{\theta}_2}(\bm{x}|\bm{z})]  
\label{eq:lossVAE}
\end{split}
\end{align}
The loss function consists of two components: a Kullback-Leibler divergence term $ D_{\textrm{KL}}\big(q_{\boldsymbol{\phi}}(\bm{z}|\bm{x}) || p_{\boldsymbol{\theta}_1}(\bm{z})\big) $ acting as a regularization term and an expected reconstruction error $ \mathbb{E}_{q_{\boldsymbol{\phi}}(\bm{z}|\bm{x})} [\log p_{\boldsymbol{\theta}_2}(\bm{x}|\bm{z})] $.
In the Gaussian case the term $D_{\textrm{KL}}$ can be integrated analytically. The expected reconstruction error, however, can not be computed directly and is therefore commonly estimated through sampling. Computing the gradient of the expectation term  $ \mathbb{E}_{q_{\boldsymbol{\phi}}(\bm{z}|\bm{x})} [\log p_{\boldsymbol{\theta}_2}(\bm{x}|\bm{z})] $ then faces a general issue since its backpropagation path includes a random node and sampling processes do not have gradients. To bypass this problem a mathematical operation, called the \textit{reparameterization trick}, is used. 
During the training phase optimization is done by jointly adapting the encoder and decoder parameters $ \boldsymbol{\phi} $ and $ \boldsymbol{\theta}_2 $. 
Due to the sampling process, the reconstruction will not be an exact representation of the input instance but rather constitute a (similar) new encoded data point within the latent space.

\section{Method}
\begin{figure}[t]
	\centering
	\def\svgwidth{0.75\columnwidth} 
	\includegraphics{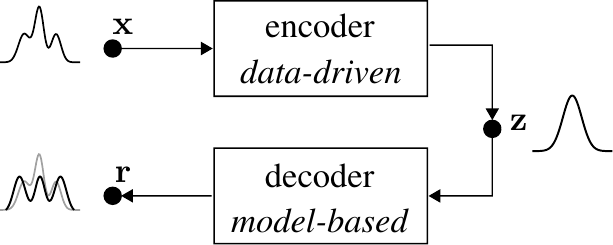}  
	\caption{The proposed architecture uses a model-based decoder, that generates an interpretable latent space. Due to the application, the common AE objective of reconstructing the input is adapted to a prediction task.}
	\label{figStructure}
\end{figure}
The proposed model's functionality is based on the objective of VAEs due to the mentioned benefits. Since the VAE's latent features are not interpretable by nature, a conceptual adaption is applied in order to attribute it with a meaningful semantic: Instead of relying solely on a data-driven learning process for the encoder-decoder structure, the decoder's inner functionality is based on expert knowledge. As illustrated in Figure \ref{figStructure}, the decoder solves a subtask of the overall problem through a model-based consideration. Its computational steps embody explicit equations with a predefined number of $\mathcal{Z}$ parametrizable variables. This \textit{descriptive computing block} uses the latent space $\bm{z} \in \mathbb{R}^\mathcal{Z}$ as input to its decoding function. Conventionally, the goal of VAEs is to reconstruct the network's input: By penalizing the prediction according to Eq.\,\ref{eq:lossVAE}, the encoder parameters $\boldsymbol{\phi}$ are adapted. In comparison to conventional VAEs, the decoder parameters $\boldsymbol{\theta}_2$ do not have to be learned. This setup forces the encoder to extract latent parameters that have a semantic meaning and enable a mathematical interpretation. Depending on the implemented functions in the decoder, the latent space takes on the corresponding semantics of the parameters in the deterministic mapping. Hence, the latent space is interpretable. The proposed architecture is referred to as the \textit{Descriptive Variational Autoencoder (DVAE)}.\\
\textbf{Encoder:}
The encoder is a conventional NN whose network parameters are optimized during training. Its architecture depends on the task complexity and application.\\
\textbf{Decoder:}
The implementation of the decoder as a descriptive computing block allows to have an interpretation of the latent space. This also holds, when minor learning tasks remain within the decoder. According to the task, it's internal computations are based on domain knowledge. 

Note, that in comparison to the VAE the assumption regarding the latent distribution of the DVAE has to be consistent with the expected distributions. As soon as the latent space has a semantic interpretation, the latent prior $p_{\boldsymbol{\theta}_1}(\bm{z})$ can not be chosen arbitrarily.

\subsection{Application}
Since this paper's application is a prediction task, the optimization objective of the DVAE is adapted: Instead of reconstructing the input, the architecture is trained to correctly predict the long-term trajectory of a vehicle driving on a highway. Vehicle dynamic has been intensively studied over the last decades and provides a lot of useful knowledge, e.\,g., that cars are non-holonomic systems. The movement patterns vehicles carry out can be easily described by specific equations when their intention is known. Detecting intentions within a traffic scenario, however, is a highly complex task and formulating a general policy is hardly possible. Therefore, it makes sense to assign intention recognition to be solved by a data-driven model. Predicting the corresponding motion, on the other hand, can be handled by a model-based approach. The goal is to combine these two aspects through the introduced approach. 
While the encoding part is designed in a conventional way (data-driven), the decoder is implemented by domain knowledge in the field of vehicle dynamic: It consists of model-based equations to compute the trajectory of a vehicle driving on a highway. During training, the predicted trajectory is compared to the actually executed trajectory and parameters of the encoder are optimized iteratively. For a mathematical consideration of the motion dynamics, a 2D vehicle fixed coordinate system is introduced for each vehicle: The x-axis runs horizontally in direction of the main vehicle motion, the y-axis is perpendicular to the longitudinal center plane and oriented to the left. (cf. Figure \ref{fig2})
\subsubsection{Encoder}
The encoder of the VAE is aligned to recent works of \cite{Alahi.2016} and \cite{Deo.2018} for data-driven motion prediction. The input is an observation sequence of a defined number of vehicles participating at the cooperative context. Each vehicle or possible neighbourhood position, respectively, is assigned to a single LSTM. The hidden states of all LSTMs, which are said to hold the encoded intentions of each vehicle, are then fed into a \textit{Feedforward Neural Network (FNN)}. The bottleneck of the network is represented by the latent space $\mathbf{z}$ with the distribution parameters $\boldsymbol{\mu}, \boldsymbol{\sigma} \in \mathbb{R}^\mathcal{Z}$ with $\mathcal{Z}=3$. After sampling, the latent parameters are denoted $ z_1 $, $ z_2 $ and $ z_3 $. The distribution assumptions commonly used in the VAE are appropriate for the setup and no further adaptions are necessary.
\subsubsection{Decoder}
The implemented decoder for the trajectory prediction consist of two decoupled computational chains that address the longitudinal (x-axis) and lateral (y-axis) movement. Given a latent representation, the \textit{descriptive decoder} predicts the 2D spatial trajectory $ \boldsymbol{\hat{C}} = [\bm{\hat{x}}, \bm{\hat{y}}] $ for the defined prediction horizon $ t_\mathrm{pred} $.

\textbf{Longitudinal Movement Prediction:}
The longitudinal trajectory of a vehicle is predicted through a constant acceleration model. Based on the fact that no initial spatial or temporal offset ($x_0,\, t_0=0$) is considered for the predictions, the distance traveled at any time step $t_i$ can be determined through
\begin{equation}
\hat{\mathrm{x}}(t_i, a_x) =  v_{0,x} t_i + 0.5 a_{x} t_i^2\, .
\label{eq:hatXpred}
\end{equation}
For computation of the spatial predictions within the prediction period, let $ \bm{t} = [t_0, t_0+\Delta t, t_0 + 2\Delta t, \dots , t_{\textrm{pred}}]^\mathrm{T}$ be the vector containing the timestamps $t_i$. The time increment $ \Delta t $ and the prediction horizon $t_\mathrm{pred}$ define the total number of prediction points $ P $. The initial longitudinal velocity $v_{0,x}$ is parameterized as the most recent velocity information before the prediction. The latent parameter $ z_1 $ is interpreted and used to provide the longitudinal acceleration $z_1=a_x$. During training, the encoder learns to provide the most probable acceleration for the latent parameter $z_1$ given the cooperative context information.

\textbf{Lateral Movement Prediction:}
Since in this work the use case is limited to highway scenarios, the vehicles are expected to perform only small steering maneuvers such as changing the lane. A simplified assumption for the common course behavior of vehicles can be done: the general movement of vehicles on highways can be approximated by a s-shaped curve. As a representative for a s-shaped function the sigmoid function is used because it provides beneficial characteristics that fortify its usability: monotony and smoothness. These two properties resemble the average behavior of movements in traffic: vehicle occupants usually prefer comfort and, hence, pursue monotone/smooth movements. However, its native characteristic does not allow to represent the needed complexity of real trajectories. To allow more flexibility for the predictions, the sigmoid function is extended by two additional scaling parameters $ \lambda $ and $ \mu $. Eq.\,\ref{eq:AdaptedSigmoid} represents the customized sigmoid function $ \sigma_c(\tau) $, which is used to approximate the lateral path of the target vehicle.

\begin{equation}
\sigma_c(\tau) = \lambda \frac{1}{(1 + e^{-\mu \tau})}
\label{eq:AdaptedSigmoid}
\end{equation}
The parameters describing the function are $ \lambda $ for scaling the amplitude, $ \mu $ for scaling the length, and $ \tau $ for selecting a evaluation position or time step, respectively.
Note, that the sigmoid function is a zero centred function. To make use of its complete s-shaped course, the vector containing all evaluation positions $\boldsymbol{\tau}$ has to be zero centred as well. Hence $\tau_i$, the i-th entry of $\boldsymbol{\tau}$, is definded as
\begin{equation}
\tau_i = t_i - 0.5 t_\mathrm{pred}\, .
\end{equation}
The resulting $\tau_i$ refers to the prediction time instance $t_i$ and the resulting value of the customized function $ \sigma_c(\tau_i) $ to the lateral vehicle position at $t_i$. 
The sign of the parameter $ \lambda $ indicates the direction of the movement and the absolute value indicates how far the vehicle moves into the corresponding lateral direction. A parameterization by zero specifies that no lateral movement is predicted. The stretching value is applied in order to define a varying smoothness of the s-shaped path. The smaller its value, the smoother the path will be. For values close to zero a linear function is approximated. This depicts the scenario of keeping up the lane.
To avoid a positional jump for the initial prediction step $ {\tau}_0 $ due to scaling, the (potential) zero-offset in the prediction has to be subtracted. The equation for the lateral trajectory, thus, results in
\begin{equation}
\hat{y}({t_i}, \lambda, \mu) = \lambda \frac{1}{(1 + e^{-\mu \tau_i})}) + (-\lambda) \frac{1}{(1 + e^{-\mu \tau_0})} \, ,
\label{eq:hatY}
\end{equation}
where $ \lambda = z_2 $ and $ \mu = \textrm{exp}(z_3) $, with $ \mu \in \mathbb{R}^+_0 $. By inverting the sign of the exponent, the sigmoid function would be mirrored around the y-axis. This flexibility is unnecessary and therefore suppressed by setting $ \mu = \textrm{exp}(z_3) $.  
\subsubsection{Backpropagation}
For using gradient based learning approaches, the backward path of the decoder has to be determined. It results in the following partial derivatives \ref{eq:8}-\ref{eq:11}.
\begin{equation}
\frac{\delta \hat{x}(t_i, a_x=z_1)}{\delta z_1} = 0.5t_i^2 \label{eq:8}
\end{equation}
\begin{align}
\frac{\delta \hat{y}(t_i, \lambda= z_2, \mu)}{\delta z_2} &=
\frac{1}{1+e^{-\mu {\tau_i}}} - \frac{1}{1+e^{-\mu {\tau}_0}} \nonumber \\
&= \sigma_c \left(
{\tau}_i\right) - \sigma_c \left({\tau}_0\right)
\end{align}
\begin{align}
\begin{split}
\frac{\delta \hat{y}(t_i, \lambda, \mu=e^{z_3})}{\delta z_3}
=& \lambda e^{z_3} \big( \sigma_c ({\tau}_i)\left(1-\sigma_c({\tau}_i)\right){\tau}_i \\
&- \sigma_c ({\tau}_0)\left(1-\sigma_c({\tau}_0)\right){\tau}_0\big)
\end{split} \label{eq:11}
\end{align}
\subsection{Validation through the Interpretable Latent Space}
\begin{figure}[]
	\footnotesize
	\includegraphics[width=0.95\columnwidth]{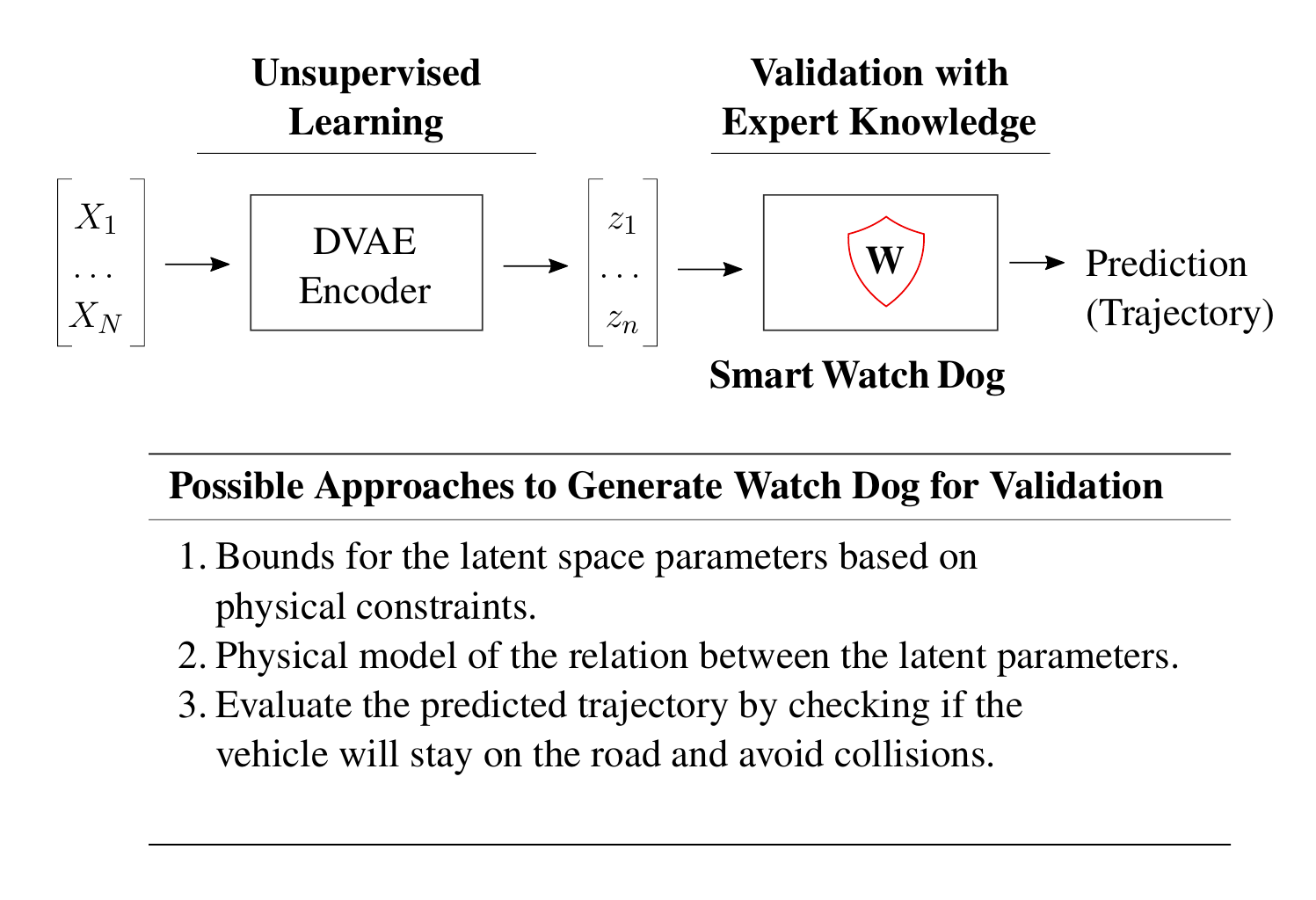}
	\caption{Expert-assisted validation based on the interpretable latent space.}
	\label{fig:MarionsWatchDog}
\end{figure}

Through the descriptive design of the decoder, the latent space allows a certain physical interpretation. Due to this interpretation, predictions of the NN can be validated by using expert knowledge regarding the latent space. Based on the physical interpretation of each parameter, application dependent rules can be introduced in order to validate the network prediction. In Figure \ref{fig:MarionsWatchDog} the general concept of the validation conditioned on the latent space is illustrated. In this work, the latent parameters that allow evaluation are $\lambda$, $\mu$ and $a_x$. These parameters define the predicted trajectory. By introducing rules for these interpretable parameters of the NN, wrong trajectory predictions can be easily detected which allows to accelerate validation. 
\section{Experiments}
The proposed DVAE is benchmarked against three different approaches for trajectory prediction. 
For training and testing the publicly available highD dataset \cite{Krajewski.} is used due to its extend of application-oriented scenarios. The dataset provides traffic observations of different German highways. Each recorded highway section is about \SI{420}{\meter} long and captures both driving directions. To force the model to conclude from traffic patterns instead of absolute position information, the global description of the highD dataset is transformed into local (vehicle centered) descriptions. The resulting data format is aligning with the problem definition explained in the following. Experiments were implemented in Python 3.7.6, using PyTorch 1.4.0, and executed in Ubuntu 19.10 on a NVIDIA GeForce RTX 2070 and 32\,GB memory.

\subsection{Problem Definition}
The experimental task is to predict the trajectory of a single target vehicle in a highway scenario, given motion observations of the target vehicle and its environment. In total, $ N=8 $ positional neighbouring road users (and the target vehicle itself) are considered in the prediction process. At each timestep the relative spatial coordinates of the surrounding road users w.\,r.\,t. the predicted agent are given. The motion information of the $ j $-th ($ j = 1, ...,N $) participating road user is represented as $ \boldsymbol{\xi}_j = [\bm{x}_{j,\textrm{rel}}, \bm{y}_{j,\textrm{rel}}, \bm{v}_{j,x,\textrm{rel}}, \bm{v}_{j,y,\textrm{rel}}] \in \mathbb{R}^{O \times 4}$, with $O$ being the total number of observed timestamps.
The input matrix of a single road user $ \boldsymbol{\xi}_j $ contains the relative distances and velocities within the observation period $ t_\textrm{obs}=\SI{3}{\second}$ up to the
current time step $ t_0 $.
During observation, the origin of the used coordinate frame is fixed within the target vehicle's origin. Thus, the observation information of the predicted agent includes only its absolute velocities $ \boldsymbol{\xi}_0 = [\bm{v}_{0,x,\textrm{abs}}, \bm{v}_{0,y,\textrm{abs}}] \in \mathbb{R}^{O \times 2}$. 
The highway traffic scenario data $\mathcal{D} = \left\lbrace (\mathbf{X}_1,\mathbf{C}_1),\dots,(\mathbf{X}_M,\mathbf{C}_M) \right\rbrace$ includes the stacked dynamic observations $\mathbf{X}_m = [\boldsymbol{\xi}_0, \boldsymbol{\xi}_1, \dots , \boldsymbol{\xi}_N] \in \mathbb{R}^{O \times (2+4N)}$ and the target trajectory $\mathbf{C}_m =\left[ \bm{x}, \bm{y} \right]\in \mathbb{R}^{P \times 2}$. By processing the input relations, the model predicts the most probable trajectory of the target vehicle $ \boldsymbol{\hat{C}} = [\bm{\hat{x}}, \bm{\hat{y}}] \in \mathbb{R}^{P\times 2}$ within the prediction period from $ t_0 $ to $ t_\textrm{pred} = \SI{5}{\second}$.

\subsection{Compared Methods}
The experiment includes a comparison of the proposed DVAE with benchmarking networks: a conventional VAE, an AE with the descriptive decoder (DeAE) and a model-based approach. 
\begin{figure*}[htp!]
	\centering
	\includegraphics[width=0.70\textwidth]{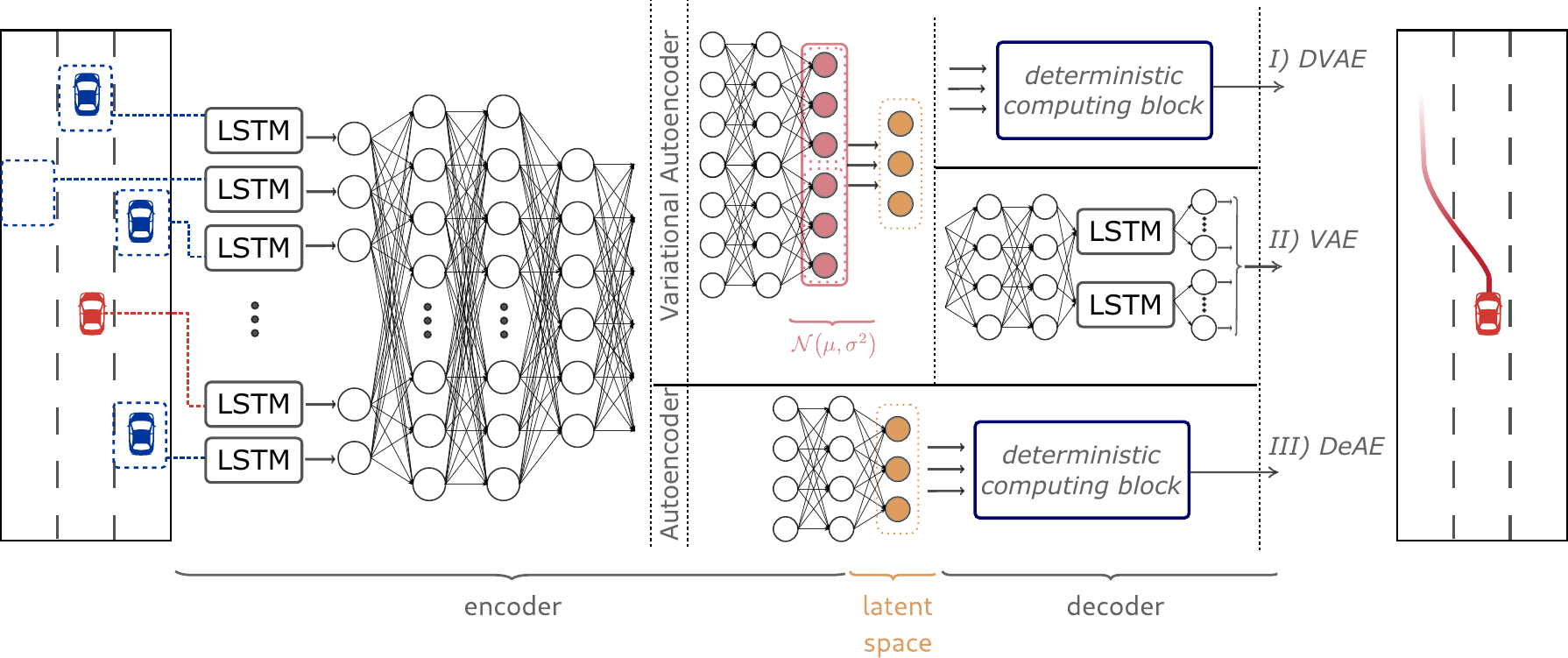} 
	\caption{Architecture of the compared methods for trajectory prediction. All networks have the same encoder setup and only their decoding parts are dissimilar. Determination of the latent space, however, naturally differs between VAEs and AEs.}
	\label{fig}
	\vspace{-10pt}
\end{figure*}
Note here, that the main scope is not to compare the approach to the currently best performing trajectory prediction networks or even outperform them. The scope here is to introduce an unsupervised prediction concept that provides interpretability, show its feasibility and evaluate the potential of its prediction performance. An improvement of the performance and comparison to other supervised trajectory planning methods is planned for future work.
The compared data-driven methods (DVAE, VAE, DeAE) share a common encoder architecture. Meaning, all layer dimensions and amount of neurons within the encoder part are equal. However, the latent space determination between VAE- and AE-based approaches differ naturally. Figure \ref{fig} depicts the data-driven setups according to the network complexity. The transpose of the observation sequence of the target vehicle $\boldsymbol{\xi}_0$ is input to a LSTM of the dimension [Input$\times$hidden$\times$output: $ 2\times8\times64 $]. Each of the other $ N $ surrounding vehicles provides its transposed input matrix $ \boldsymbol{\xi}_j^\mathrm{T} \in \mathbb{R}^{4\times O} $ to the corresponding LSTM of the dimension [Input$\times$hidden$\times$output: $ 4 \times 16 \times64$]. In order to extract the cooperative context information, the hidden states of all LSTMs are passed onto a FNN with four layers of the decreasing dimensions $ 136-64-64-18 $. From this point onwards, the implementations differ:
\begin{itemize}
	\item \textbf{I) Descriptive Variational Autoencoder (DVAE):} The values of the last layer's 18 neurons are transformed into a space with six values, representing the distribution parameters $\bm{\mu}, \bm{\sigma} \in\mathbb{R}^3$. Through sampling from this distribution, an element of the desired latent space $z\in\mathbb{R}^3$ is generated. The decoder is the descriptive computing block. In total, the DVAE provides $30.020$ learnable parameters.
	\item \textbf{II) Variational Autoencoder (VAE):} The values of the last layer's 18 neurons are transformed into a space with six values, representing the distribution parameters $\bm{\mu}, \bm{\sigma} \in\mathbb{R}^3$. Through sampling from this distribution, an element of the desired latent space $z\in\mathbb{R}^3$ is generated. The latent space is forwarded through another two layers with 16 and then 64 nodes. The resulting output serves as input to two separate LSTMs [Input$\times$hidden$\times$output: $64\times125\times125$], one for the longitudinal trajectory and the other one for the lateral trajectory. After each LSTM another layer is appended, supplying 125 nodes each, in order to produce the desired output dimension $P$. In total, the VAE provides $214.613$ learnable parameters.
	\item \textbf{III) Descriptive Autoencoder (DeAE):} The three dimensional latent space is computed from the 18 neurons. The decoder is the descriptive computing block. In total, the DeAE provides $29.963$ learnable parameters.
\end{itemize}
The model-based approach is implemented as follows.
\begin{itemize}
	\item \textbf{Constant Velocity (CV) model:} The CV model predicts the target vehicle's trajectory according to Eq.\,\ref{eq:CVx} and \ref{eq:CVy}. Within the prediction horizon $t_\textrm{pred}$ the position is computed using the constant decoupled velocities $v_x$ and $v_y$. The latest available velocity information is used and initial offset ($x_0, y_0$) is set to zero. 
	\begin{align}
	\hat{x}_\textrm{CV}(t_i) = x_0 + v_{0, x} t_i \label{eq:CVx} \\
	\hat{y}_\textrm{CV}(t_i) = y_0 + v_{0, y} t_i \label{eq:CVy}
	\end{align} 
\end{itemize}

\subsection{Classification}
An simple application of the unsupervised technique is to use the interpretable latent space for classifications. Based on the physical meaning of the parameters in the latent space a classifier can be implemented using simple rules.
Due to the inherent semantic of the parameters, a mathematical interpretation is practicable. The classifier predicts whether the target vehicle will carry out a lane change and distinguishes the maneuver classes $ C $: Keep Lane (KL), Lane Change Left (LL) and Lane Change Right (LR). Therefore, the thresholds $ t_\lambda  $ and $ t_\mu $ on the parameters defining the latent movement are introduced and the following logic is applied:

\textbf{IF} $ (\mu < t_\mu) $ \textbf{OR} $ (\textrm{abs}(\lambda) < t_\lambda) $ \textbf{then} $ C \leftarrow \textrm{KL} $

\textbf{ELSE IF} $ (\lambda > t_\lambda) $ \textbf{then} $ C \leftarrow \textrm{LL} $

\textbf{ELSE }$ C \leftarrow \textrm{LR} $ \, . \\
\subsection{Implementation Details}
The networks are trained using the Stochastic Gradient Descent (SGD) algorithm and a learning rate of $ \alpha = 0.001 $. For training and testing, the first nine subdatasets of the original highD dataset are preprocessed. The preprocessed dataset holds $M = 130\,000$ samples, of which 2/3 are used for training and 1/3 for testing. Maneuver types are equally distributed. In total $ E = 5 $ training epochs are carried out. For the testing phase, the sampling step of the DVAE as well as of the traditional VAE implementation are turned off and the computed mean is used instead. According to the data frequency of the highD dataset the time increment is set to $ \Delta t = \SI{1/25}{\hertz} =  \SI{0.04}{\second}$ and, hence, $O=75$ and $P=125$. For classification the thresholds are set to $ t_\lambda = 0.85 $ and $t_\mu = 0.25$. Both values were gathered through evaluating the parameter distribution resulting from a curve-fitting of the target trajectories w.\,r.\,t. Eq.\,\ref{eq:hatY}. By extracting the optimal parameterization for different instances of the training dataset, the true distribution $ p(\bm{z}) $ over latent space parameters can be approximated.
\section{Results}
This section mainly addresses a performance evaluation of the predictions and based on this, stresses the usefulness of interpretability of the latent space.
\subsection{Comparison of Methods}
For evaluation of the trajectory prediction, the \textit{Empirical Cumulative Distribution Function (ECDF)} is used. The ECDF is an empirical estimation of the cumulative distribution function and defined as
\begin{align}
f(e) =  \frac{1}{n} \sum_{i=1}^{n} \bm{1}_{(e_{y_i} \leq e)} \, ,
\end{align}
\begin{figure}
	\centering
	\def\svgwidth{0.97\columnwidth} \footnotesize 
	\includegraphics[width=0.97\columnwidth]{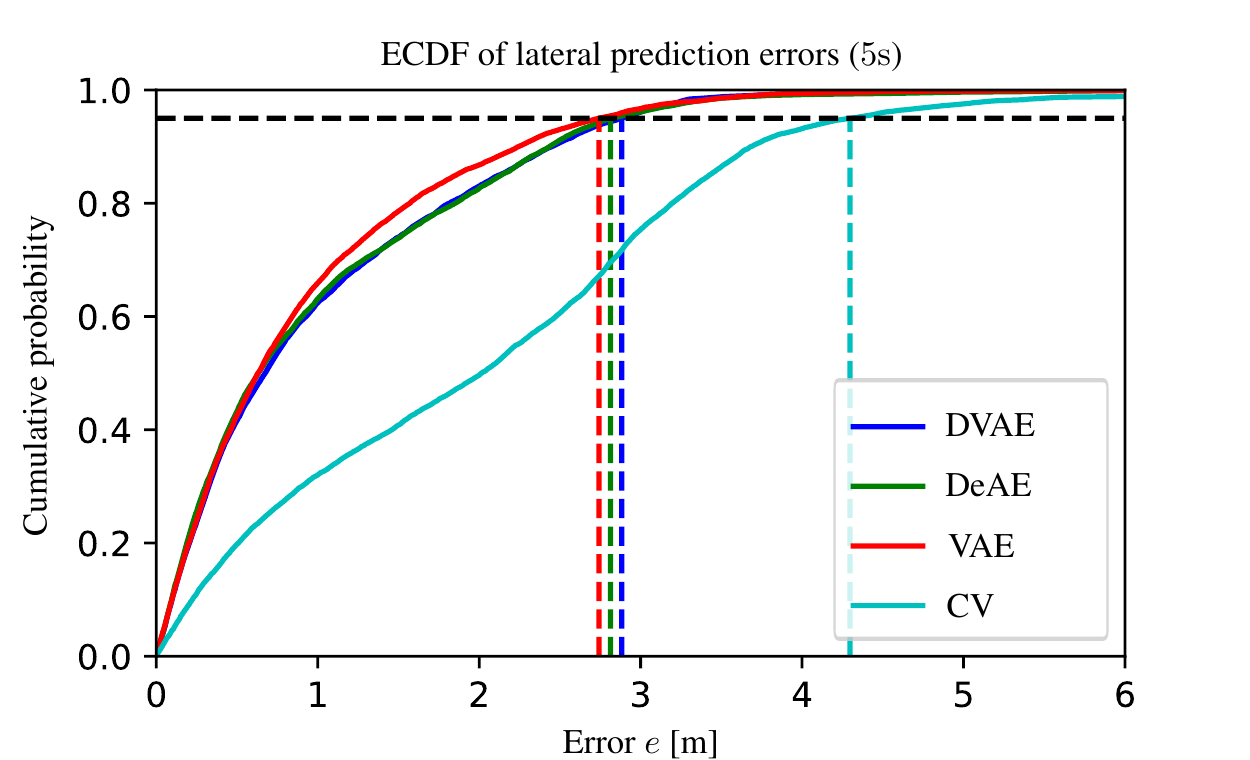}  
	\caption{The lateral prediction performance, evaluated through the ECDF metric, of the descriptive methods is similar to the prediction accuracy of the conventional VAE.}%
	\label{fig:ecdf}
	\vspace{-5pt}
\end{figure}
\begin{table}[htp!]
	\centering
	\begin{tabular}{cc|c|c|c}
		&&\multicolumn{3}{c}{\textbf{Predicted maneuver}}\\
		&\shortstack{DVAE / \\DeAE} & LL & KL & LR \tstrut \bstrut \\ \hline   \tstrut
		\parbox[t][][l]{5mm}{\multirow{3}{*}{\rotatebox[origin=c]{90}{\textbf{\shortstack{Target\\ maneuver}}}}} 
		&LL & 0.71 / 0.87 & 0.28 / 0.11 & 0.01 / 0.02 \\ \tstrut \bstrut
		&KL & 0.10 / 0.26 & 0.79 / 0.58 & 0.11 / 0.16 \\ \tstrut \bstrut
		&LR & 0.00 / 0.02 & 0.21 / 0.15 & 0.79 / 0.83 \\ \bstrut
	\end{tabular}
	\caption{Combined confusion matrix for the classification performance of the DVAE and DeAE.}
	\label{table2}
	\vspace{-15pt}
\end{table}
where $n$ is the total number of instances and $\bm{1}_{(e_{y_i} \leq e)} $ the indicator of the event $e_{y_i} \leq e$, that the lateral prediction error $e_{y_i}$ is less than or equal to a fixed error value $e$. This metric is commonly applied to asses positioning errors. When analyzing prediction performance w.\,r.\,t. reference predictions, it makes sense to investigate the (empirical) distribution of the prediction errors. Validation is purposely restricted to the lateral motion. Generally, longitudinal motion prediction is less complex and less related to the cooperative context. Therefore it has no priority in this work. Figure \ref{fig:ecdf} shows the ECDF of the lateral trajectory prediction errors, caused by each implementation.
The error $e$ on the x-axis indicates the Euclidean distance between the prediction and the reference trajectory. For evaluation, a percentile of 95\,\si{\percent} is marked. As it can be seen, the long-term trajectory prediction of the implemented ML models are almost 50\,\si{\percent} more accurate than the CV predictions. The slope of the CV model's CDF is much more shallow and, hence, implies a much lower overall confidence in its forecasts. 
More interesting is the fact that the performance of all ML models is almost the same: the prediction error is $ \sim \SI{2.85}{\meter}$ or smaller at 95\,\si{\percent} of the predictions. The VAE does perform slightly better than the descriptive models but also provides much more degrees of freedom; 300 additional nodes and two complex LSTMs. Comparing the descriptive methods by the ECDF, the performance of the implemented DeAE and DVAE is similar and non of them is clearly superior.
\begin{figure}[!htb]
	\centering
	\includegraphics[width=0.75\columnwidth]{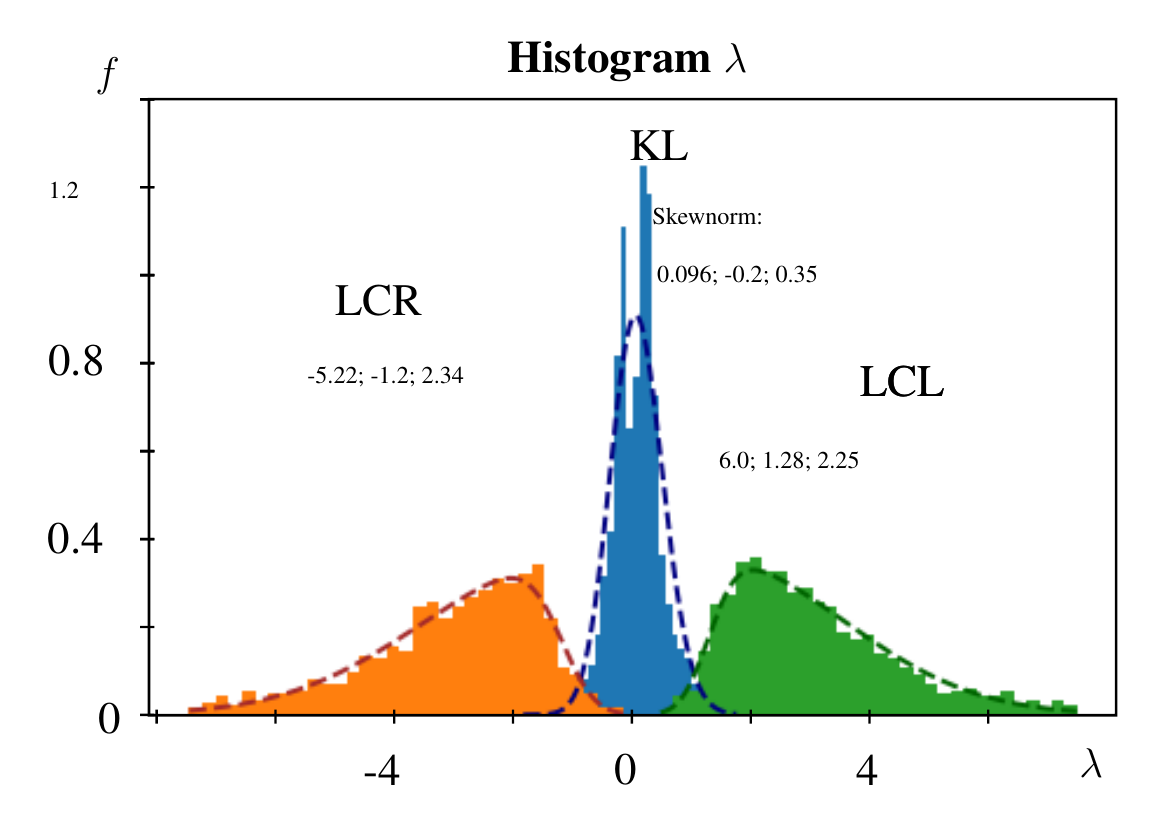}  
	\caption{Reference histogram of the latent parameter $\lambda$ with a skew normal distribution approximation (shape; location; scale).}
	\label{fig5}
\end{figure} 
\begin{figure}[htp!]
	\centering
	\includegraphics[width=0.9\columnwidth]{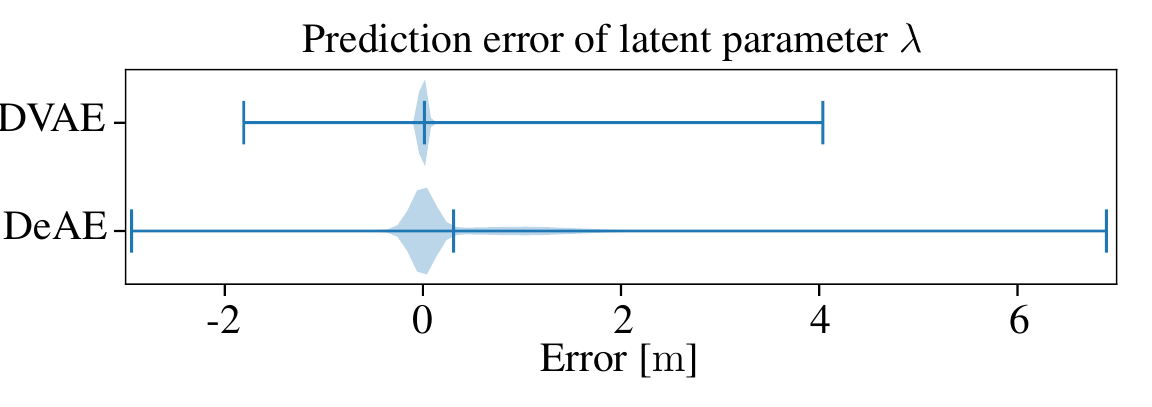}  
	\caption{Mean, distribution and range of prediction errors of the parameter $\lambda$ according to the used network.}
	\label{pred_error_plot}
	\vspace{-8pt}
\end{figure}
When examining their performance in the simple threshold-based classification, an interesting result arises. While both models provide an average prediction accuracy of 76\,\si{\percent}, the DeAE is noticeable worse in detecting KL-maneuvers with an accuracy of 58\,\si{\percent} as can be seen in the confusion matrix in Table\,\ref{table2}.
This issue is due to the distinct learned latent parameter distributions with respect to the classification boundaries. Figure \ref{fig5} illustrates the reference distributions of the parameter $ \lambda $. It is computed by applying a curve-fitting method to get the best suited parameterization for $\lambda$ and $\mu$. This is done in a non-causal fashion, which is why it can't be used in prediction systems. 
The reference data shows that the maneuver boundaries are at the threshold of $t_\lambda = \pm 0.85$, as already used for the classification. In order to achieve a proper maneuver classification, the intersection of the distinct hypotheses (LCR, KL, LCL) are required to be learned correctly. Hence, if the parameter range for which the parameter $\lambda$ indicates that the vehicle keeps the lane is learned inaccurately (e.\,g. broader than it actually is) the resulting classification performance is excessively non-uniform among the classes. This is the case for the DeAE. Since the DVAE is based on a variational loss function, that explicitly addresses the latent parameter distribution, the hypotheses intersections can be learned accurately easier.
To stress this aspect and, accordingly, the superiority of the DVAE, the absolute prediction errors of the parameter $\lambda$ resulting from both models are evaluated in Figure \ref{pred_error_plot}. The classification accuracy is depending on the prediction error for $\lambda$. The DeAE's prediction of the parameter $\lambda$ has a larger maximal error, a greater variance in the error distribution and is biased. Hence, the DVAE's prediction for parameter $\lambda$ is more accurate in general which is an important fact regarding a prediction validation based on the interpretable latent space as explained in the following.
Note, that both models did not have access to the reference data of Figure \ref{fig5}\ i.\,e., the DVAE and DeAE are trained completely unsupervised. According to the goal, that the model should be able to learn an interpretable latent space in an unsupervised manner, it was not intended to preprocess the data in order to then use the resulting physical values for a supervised learning. Such an unsupervised approach holds the advantage of being automatable, even for more complex decoder structures.
\subsection{Expert-assisted Trajectory Validation with the \\Interpretable Latent Space}
The minimal performance benefit of the VAE comes at the high cost of missing interpretability. While the VAE's prediction could end up in an arbitrary or unrealistic path, this is no danger for the descriptive models. The predicted trajectories of these models result directly from the latent space parameters. Solely these parameters have to be checked for validity, which is far more straight forward than evaluating a set of prediction points. According to the physical interpretation of the latent space, implausible values can be detected by checking if their values lead to an unfeasible driving dynamics or exceed a plausible range.
An explicit example would be to evaluate the condition $||\lambda||<8$: Larger $\lambda$ values are not realistic for normal driving situations on highways and indicate a false prediction. As can be taken from the Figure \ref{fig5} the common range for the parameter $\lambda$ is approximately between $[-7.5, 7.5]$. If this range is exceeded, the trajectory prediction will most likely be wrong. Such kind of information can be used to design the "Smart Watch Dog" from Figure \ref{fig:MarionsWatchDog}.
\section{Conclusion}
The proposed Descriptive Variational Autoencoder (DVAE) is a partially interpretable vehicle trajectory prediction model that is based on the concept of VAEs but uses expert knowledge within its decoder part. The DVAE comes along with two main advantages: Firstly, the dataset does not require an expensive labeling process, hence, the training is completely unsupervised. Secondly, the resulting latent space is interpretable, which allows prediction validation by rules in the latent space.
Evaluation of the proposed network shows, that DVAEs have a similar prediction accuracy as their non-interpretable counterparts. This underlines the high potential of DVAE since the provided interpretability enables systematic testing and contributes towards the acceptance of ML for crucial task solving as predictions can be evaluated by their validity. Physically impossible or unlikely feature values can be detected and declared as untrustworthy. 
Although the proposed DVAE provides interpretability, yet, its prediction accuracy is not inferior to a baseline VAE, that even provides significantly more degrees of freedom.

In future works, the goal is to introduce more sophisticated motion calculations within the decoder part to improve prediction performance. Furthermore, a more general decoder design is necessary to enable the network's usage for all traffic situations and not to be limited to highway-alike scenarios. A comparison with state of the art trajectory prediction networks on multiple datasets is planned.

\
{\small
	\bibliographystyle{ieee_fullname}
	\bibliography{ref/ref_aaai.bib}
}

\end{document}